\documentclass[runningheads]{llncs}
\usepackage{graphicx}
\usepackage{comment}
\usepackage{amsmath,amssymb} 
\usepackage{color}
\usepackage{multirow}
\usepackage{floatrow}
\newfloatcommand{capbtabbox}{table}[][\FBwidth]
\usepackage{bbm}

\makeatletter
\newcommand{\printfnsymbol}[1]{%
  \textsuperscript{\@fnsymbol{#1}}%
}
\makeatother

\begin{document}
\pagestyle{headings}
\mainmatter

\title{Pixel-BERT: Aligning Image Pixels with Text by Deep Multi-Modal Transformers} 

\titlerunning{Pixel-BERT}
%
\author{Zhicheng Huang  \inst{1}\thanks{Equal contribution. This work was conducted when Zhicheng Huang and Zhaoyang Zeng were research interns at Microsoft Research.} \and
Zhaoyang Zeng \inst{2}\printfnsymbol{1}  \and
Bei Liu \inst{3} \and
Dongmei Fu \inst{1} \and
Jianlong Fu \inst{3}
}
\authorrunning{Z. Huang et al.}
%
\institute{University of Science and Technology Beijing \\
\email{zhicheng.huang@xs.ustb.edu.cn, fdm\_ustb@ustb.edu.cn} \and
Sun Yat-sen University \\
\email{zengzhy5@mail2.sysu.edu.cn} \and
Microsoft Research \\
\email{\{bei.liu, jianf\}@microsoft.com}}
\maketitle

\begin{abstract}
We propose Pixel-BERT to align image pixels with text by deep multi-modal transformers that jointly learn visual and language embedding in a unified end-to-end framework. 
We aim to build a more accurate and thorough connection between image pixels and language semantics directly from image and sentence pairs instead of using region-based image features as the most recent vision and language tasks.
Our Pixel-BERT which aligns semantic connection in pixel and text level solves the limitation of task-specific visual representation for vision and language tasks.
It also relieves the cost of bounding box annotations and overcomes the imbalance between semantic labels in visual task and language semantic. 
To provide a better representation for down-stream tasks, we pre-train a universal end-to-end model with image and sentence pairs from Visual Genome dataset and MS-COCO dataset. 
We propose to use a random pixel sampling mechanism to enhance the robustness of visual representation and to apply the Masked Language Model and Image-Text Matching as pre-training tasks. 
Extensive experiments on downstream tasks with our pre-trained model show that our approach achieves state-of-the-art results in downstream tasks, including Visual Question Answering (VQA), image-text retrieval, Natural Language for Visual Reasoning for Real (NLVR$^2$). Particularly, we boost the performance of a single model in VQA task by 2.17 points compared with SOTA under fair comparison.

\end{abstract}


\begin{figure}[t]
\centering
\includegraphics[width=1.0\linewidth]{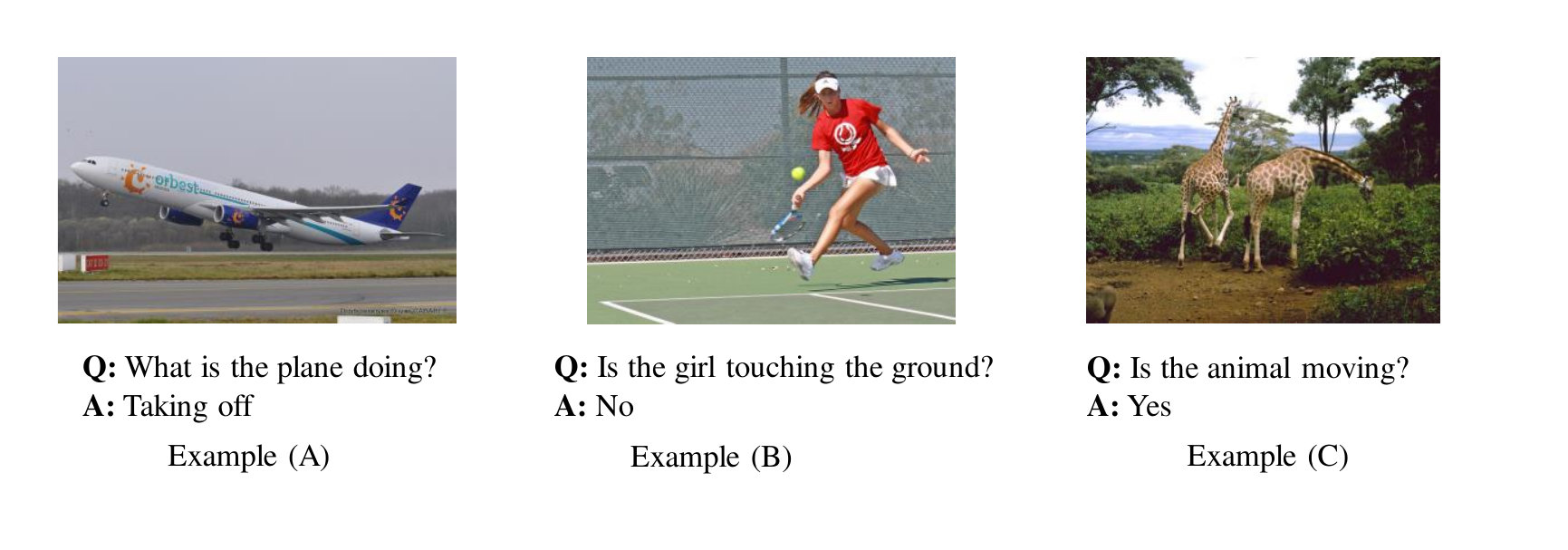}
\caption{Examples of images, questions (Q) and answers (A) in VQA2.0 dataset. In these cases, region-based visual features cannot well handle the questions.}
\label{fig:example}
\end{figure}

\section{Introduction}

With the success of self-supervised learning applied for representation learning in natural language process field \cite{vaswani2017attention,devlin2018bert}, recent research has addressed the task of cross-modality learning, especially in vision and language \cite{chen2019uniter,li2019unicoder,li2019visualbert,lu2019vilbert,qi2020imagebert,su2019vl,tan2019lxmert,zhou2019unified}, in a similar self-supervised learning way. Pre-training is widely used in those works to provide a strong representation for both vision and language in cross-modality domain. Most of them utilize BERT-based language feature and region-based visual feature as the input for joint embedding learning in pre-trained models. 

The semantic gap between different modalities has always been treated as one of the most important challenges in cross-modality research. Early works in vision and language, such as Visual Question Answering (VQA) \cite{Antol_2015_ICCV} and image captioning \cite{Vinyals_2015_CVPR}, utilize CNN features directly extracted from a pre-trained model on image classification task. Later on, with the introduction of Visual Genome Dataset \cite{krishna2017visual} and the proposal of Bottom-Up and Top-Down Attention model \cite{Anderson_2018_CVPR}, most recent vision and language methods, including the pre-trained models mentioned above, utilize region-based visual features extracted from object detection model (e.g., Faster R-CNN) for better performance. 
However, region-based visual feature extractors are designed for specific visual tasks (e.g. object detection), and this will cause an information gap with language understanding.
Some important factors of visual information are lost, such as shapes of objects, spatial relations between objects with overlap, etc. 
Moreover, the feature representation capability is limited to the given categories of such task-specific model while visual information of much broader semantics, such as scene and sentiment, are lost in the object detection model.
We show some examples that region-based visual features cannot well handle in Fig. \ref{fig:example}.
In Example (A), it is difficult for object detection models to obtain the status of the plane. For Example (B), even though we can detect the ``girl'' and ``ground'', since there is overlap between their regions, it will be even hard for further fusion embedding models to judge the actual spatial relation given their bounding boxes. Similarly in Example (C), with only visual features of ``giraffe'', it is difficult to infer the status of the animals.

Existing methods that use region-based visual features and language embedding as input of Transformer for cross-modality joint learning are limited to the visual semantics represented by the visual features.
Thus, we step out of the bounding box to make the full power of visual information in images for vision and language learning. 
We propose \textbf{Pixel-BERT} that learns to align image pixels with text to build a more thorough semantic embedding between visual and textual information. 
Pixel-BERT consists of three parts: a fully convolutional neural network (CNN) that takes image pixels as input for visual embedding learning, a word-level token embedding based on BERT, and multi-modal transformers for jointly learning of visual and language embedding.

To learn a universal representation for most vision and language tasks, we first pre-train our model with image-sentence pair dataset similar to other cross-modality pre-training methods \cite{chen2019uniter,li2019unicoder,li2019visualbert,lu2019vilbert,qi2020imagebert,su2019vl,tan2019lxmert,zhou2019unified}. Two pre-training tasks and one pre-training mechanism are used in our pre-training procedure. For language, we follow other pre-training works \cite{chen2019uniter,su2019vl,tan2019lxmert} to use Masked Language Modeling (MLM) for the prediction of masked tokens with surrounding text and image. For vision, we propose a random pixel sampling mechanism to make up for the difficulty of predicting pixel-level features. The random pixel sampling mechanism improves the robustness of visual feature learning and overcomes the problem of overfitting. For vision and language interaction, we follow \cite{chen2019uniter} to apply Image-Text Matching (ITM) to classify whether an image and sentence pair is matched or not. 

The contributions of this paper are summarized as follows:
\begin{itemize}
\item We propose Pixel-BERT that consists of a CNN-based visual encoder and deep multi-modal transformers to jointly learn visual and language embedding. We are the first to consider to align vision and language semantics in pixel and text level using self-supervised learning.
\item We use our model in pre-training manner and propose a random pixel sampling mechanism to improve the robustness of visual representation learning.
\item Extensive experiments demonstrate the effectiveness of our approach by achieving state-of-the-art performances in various tasks, including VQA, Image-Text Retrieval and NLVR$^2$. In particular, our approach improves the single model's performance of VQA by 2.17 points compared with previous SOTA \cite{chen2019uniter} in fair comparison and even higher than its larger model.
\end{itemize}

\section{Related Works}

\subsection{Pre-training Mechanism} 
For vision and language tasks, a better understanding of semantics is important to get better joint representation. For visual-content understanding, several backbone models~\cite{he2016deep,szegedy2015going,xie2017aggregated} have been proposed for pure visual understanding, which have shown their effectiveness on large datasets~\cite{deng2009imagenet}. Pioneering work~\cite{ren2015faster} also shows the generalizability of pre-trained backbone models by fine-tuning them on different downstream tasks. In terms of language understanding, we have witnessed rapid progress towards building a universal backbone model with large-scale contextualized pre-training ~\cite{conneau2019cross,devlin2018bert,radford2018improving} in recent years. They improved the performances on various tasks to significant levels. For cross-modality research, many methods~\cite{chen2019uniter,li2019unicoder,li2019visualbert,lu2019vilbert,qi2020imagebert,su2019vl,sun2019videobert,tan2019lxmert,zhou2019unified} have been proposed recently. They focus on leaning the visual and sentence dense connection between different modalities. Existing methods can be clustered into two groups based on their network structure. Some works~\cite{lu2019vilbert,tan2019lxmert} utilize two-stream neural networks based on the Transformer~\cite{vaswani2017attention}. The two-stream neural networks process visual and language information respectively and fuse them afterward by another Transformer layer. On the other hand, there are some methods~\cite{alberti2019fusion,chen2019uniter,li2019unicoder,qi2020imagebert,su2019vl} apply single-stream neural network. They use BERT~\cite{devlin2018bert} to learn a bi-directional joint distribution over the detection bounding box feature and sentence embedding feature. The differences among them are the training method, loss function, and datasets. 
Pixel-BERT is categorized into second group while our way of visual embedding is different from all those methods.

\subsection{Visual Feature Embedding in Vision and Language Tasks}
Cross-modality tasks, such as VQA~\cite{balanced_vqa_v2}, image captioning~\cite{xu2015show}, require understanding of both sentence and visual semantics. Early method~\cite{ben2017mutan} directly adopts CNN features extracted from pre-trained classification model as visual representation. Later on, with the introduction of Visual Genome Dataset~\cite{krishna2017visual} and the proposal of Bottom-Up and Top-Down Attention model~\cite{Anderson_2018_CVPR}, most recent research related to vision and language utilize region-based visual features extracted from object detection models (e.g., Faster R-CNN~\cite{ren2015faster}) for better performance~\cite{chen2019uniter,kim2018bilinear,qi2020imagebert,tan2019lxmert}. With those methods, the semantic of visual features is limited by Visual Genome detection categories. While the language domain contains much more semantic information. One key difference between our model and the other methods is the visual semantic embedding methods. Instead of using detection bounding box features as the visual semantic representation for visual and language embedding learning, we combine the visual encoder representation learning network into one framework and input the source image as the visual input. We use this model for cross-modality pre-training for learning richer visual semantics.

\begin{figure}[t]
\centering
\includegraphics[width=1.0\linewidth]{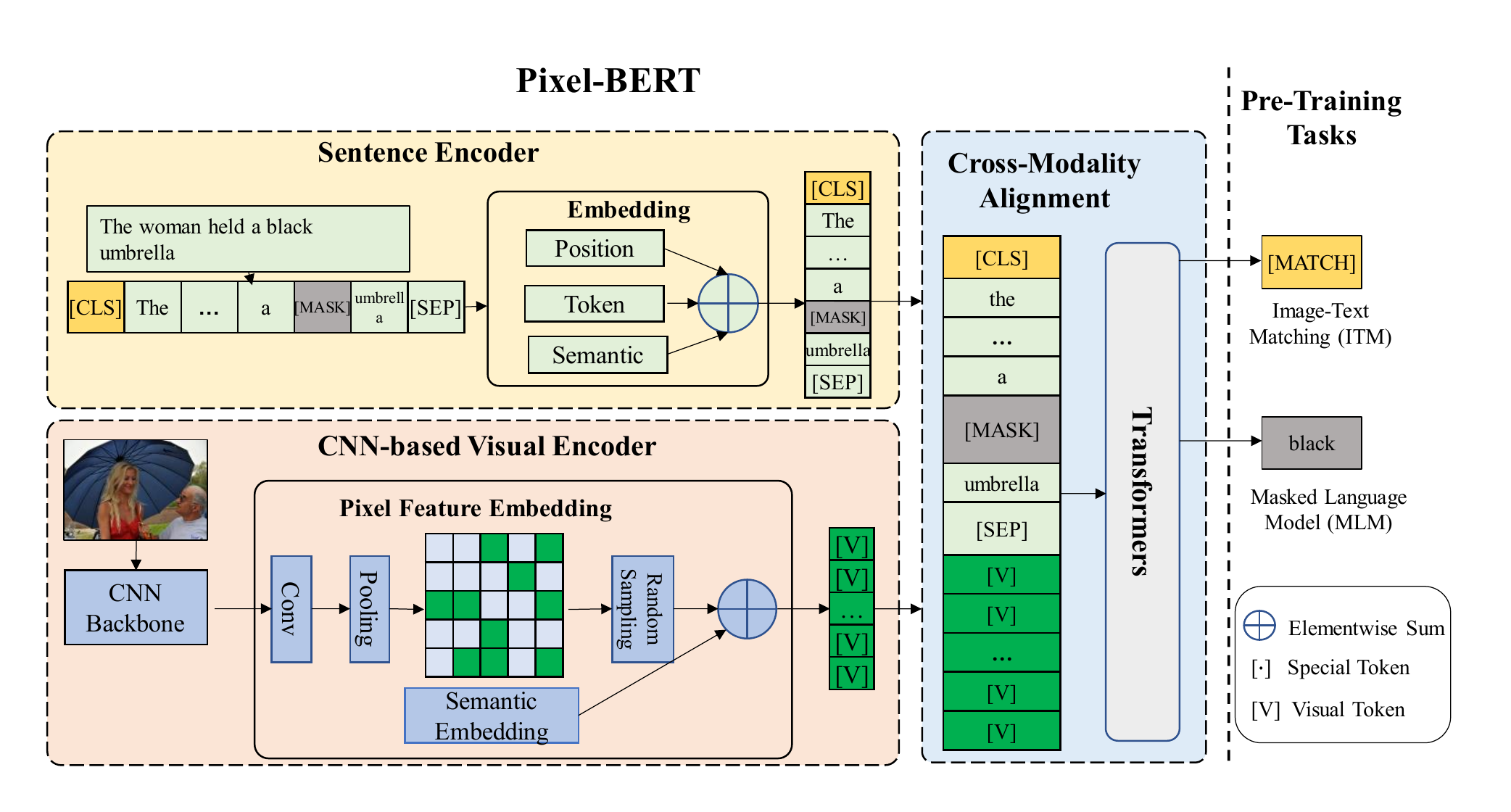}
\caption{Pixel-BERT: The model contains a visual feature embedding module, a sentence feature embedding module, and a cross-modality alignment module. Pixel-BERT takes image-sentence pairs as input, and outputs the attention features of each input element. Images are passed into a pixel feature embedding module pixel by pixel and sentences are fed into a sentence feature embedding module token by token. The model can be pre-trained by MLM and ITM tasks, and can be flexibly applied to downstream tasks (e.g. VQA, retrieval, etc).}

\label{fig:framework1}
\end{figure}

\section{Approach}
The overall architecture of our proposed Pixel-BERT, an end-to-end framework with CNN-based visual encoder and cross-modal Transformers for visual and language embedding learning, is illustrated in Fig. \ref{fig:framework1}. Image-sentence pairs are taken as input to produce joint embedding features. The whole network can be end-to-end pre-trained by MLM and ITM tasks, and is suitable to be applied to downstream tasks.

In this section, we will first revisit the Transformer model in Sec.\ref{revisit}, explain our model architecture of Pixel-BERT in detail in Sec.\ref{architecture} and pre-training procedure in Sec.\ref{pretrain}.



\subsection{Revisit Transformer}
\label{revisit}
Pixel-BERT adopts the BERT~\cite{devlin2018bert} as cross-modality alignment module. BERT is a multi-layer bidirectional Transformer encoder, which is able to model the dependency of all input elements. Before introducing our Pixel-BERT, we first revisit the architecture of Transformer. 



The two key operations in the basic Transformer module are self-attention and feed-forward. Given the input $X \in \mathbb{R} ^ {n\times d}$, where $n$ is the element number and $d$ indicates the feature dimension, we first get query $Q$, key $K$ and value $V$ from the input by
\begin{equation}
    Q = \mathcal{W}_{q}X, K= \mathcal{W}_{k}X, V = \mathcal{W}_{v}X,
\end{equation}
where $\mathcal{W}_{q}$, $\mathcal{W}_{k}$ and $\mathcal{W}_{v}$ are corresponding weight matrices.
We compute the attention output $X_{att}$ by
\begin{equation} \label{q_7}
\begin{aligned}
	&\mathcal{A}=\textnormal{softmax}(\dfrac{QK ^T}{\sqrt{d}}), \\
	&X_{att} = \mathcal{A}V,
\end{aligned}
\end{equation}
where $\mathcal{A}$ indicates the self-attention weight of each input element. The output is calculated by a feed-forward network as follows:
\begin{equation} \label{q_9}
X_{out} = \textnormal{FFN}(X_{att}),
\end{equation}
where $\textnormal{FFN}$ consists of a group of fully-connected layers with ReLU activation function as in \cite{vaswani2017attention}. Above operations build dense connections among all input elements, including each element with itself.


In cross-modality tasks, the input elements come from visual and language domains. We propose to build dense connections among both intra-domain (i.e. image-image, sentence-sentence) and inter-domain (i.e. image-sentence) by Transformer, which will be explained in detail in Sec. \ref{approach_cm}.

 
\subsection{Model Architecture}
\label{architecture}
For vision and language tasks, we have two types of inputs from different modalities.
Natural language is usually in the form of a sentence which can be split into a sequence of words. 
We follow \cite{chen2019uniter,li2019unicoder,lu2019vilbert,qi2020imagebert,su2019vl,tan2019lxmert,zhou2019unified} to first tokenize each word in the sentence, and embed each token into a vector.
The input of the visual domain is usually an image. Most recent methods represent the visual input by extracting region-based features with object detection model like Faster R-CNN\cite{Anderson_2018_CVPR}.
However, such region-based visual feature extractor is designed for specific visual tasks (i.e., object detection), which will lead to an information gap with language understanding.
Specifically, a bounding box is in the shape of a rectangle, which may include noisy background and miss the shape and spatial relation information. 
Besides, the feature representation capability is limited by the provided categories of such task-specific model.
Moreover, visual information about broader semantics such as scene and sentiment is also lost in the object detection model.
To fully utilize visual information of the original image, we propose an end-to-end framework for vision and language tasks by learning visual embedding from pixels, named Pixel-BERT. 

\subsubsection{Sentence Feature Embedding}
We follow BERT \cite{devlin2018bert} to encode the language information of a sentence. Given a sentence as input, we first split it into a sequence of words, and use WordPiece to tokenize each word into token. We then adopt an embedding matrix to embed each token into a vector. 
Here we use $\textbf{w}=\{w_{1},w_{2},...,w_{n}\}\in \mathbb{R}^{d}$ to represent the embedded sequence, where $n$ indicates the sequence length, and $d$ is the embedding dimension. 
We follow other BERT-based language methods and add the positional embedding to encode the position information. 
The final language representation of the sentence is  \{$\hat{w}_1, \hat{w}_2, \cdots, \hat{w}_n$\}. For each of the representation at position $i$, it is calculated by 
\begin{equation}
    \hat{w}_i=LayerNorm(w_i + p_i + s_w),
\end{equation}
where $p_i$ indicates the embedding vector at position $i$, $s_w$ is a semantic embedding vector and $LayerNorm$ is a normalization function described in \cite{ba2016layer}. Since the summation of position and semantic embedding is the mathematic equivalence to one embedding, we will omit the $s_w$ term in our implementation.

\subsubsection{Image Feature Embedding}
Most recent vision and language methods follow Bottom-Up and Top-Down Attention\cite{Anderson_2018_CVPR} to extract visual features by Faster R-CNN \cite{ren2015faster} trained on Visual Genome dataset. The detector extracts region features by first detecting regions under pre-defined categories, and then uses the features before the final classifier as the output. The representation ability of such extracted features will be limited to the detection categories.
To overcome the limitation of task-specific categories, shapes and borders, we learn from pixels to represent an image instead of using bounding boxes. The pixel features are learned by a CNN visual backbone such as ResNet\cite{he2016deep}. Given an input image $I$, we first use CNN backbone to extract its feature, then flat the feature along the spatial dimension. We denote the flatten feature as $\textbf{v}=\{v_{1},v_{2},...,v_{k}\}\in \mathbb{R}^{d}$, where $k$ indicates the number of feature pixels. 
The visual embedding feature $\{\hat{v}_{1},\hat{v}_{2},...,\hat{v}_{k}\}$ can be computed by
\begin{equation}
    \hat{v}_i=v_{i}+s_{v},
\end{equation}
where $s_v$ is a semantic embedding vector to distinguish the difference with language embedding. Since all pixels share the same $s_v$, this embedding vector can be considered as a bias term to be combined with the CNN backbone. In our implementation, we adopt ResNet or ResNeXt as backbone, and add a $2\times 2$ max pooling layer to reduce the spatial dimension of visual feature maps. The spatial size of input image $I$ will be down-sampled by 64 times in total.

\subsubsection{Cross-Modality Module} 
\label{approach_cm}

We adopt Transformer to learn cross-modality attention between image pixels and language tokens. After obtaining sentence embedding vectors and pixel features, we combine all vectors to construct the input sequence. We also adding two special tokens [CLS] and [SEP] for learning joint classification feature and specifying token length, respectively. The final input sequence to the joint-learning Transformer is formulated as 
\begin{equation}
    \{[\textnormal{CLS]},\hat{w}_1,\hat{w}_2,\cdots,\hat{w}_n,\textnormal{[SEP]},\hat{v}_1,\hat{v}_2,\cdots,\hat{v}_k\}.
\end{equation}
The CNN backbone for visual representation learning and the Transformer for language representation learning is combined into a single model, which is end-to-end trainable. When we apply learning supervision on the output of Transformer, the gradient can backward to the CNN backbone, and thus the learned visual features will be more suitable to the target task learning by breaking the domain gap between visual and sentence domain.


\subsection{Pre-Training}
\label{pretrain}
In order to learn a universal visual and sentence representation for vision and language related tasks, we apply the self-supervised method to pre-train a model on a large aggregated dataset. We follow \cite{chen2019uniter,li2019unicoder,lu2019vilbert,qi2020imagebert,su2019vl,tan2019lxmert,zhou2019unified} to conduct two pre-training tasks, including 
 Masked Language Modeling (MLM) and Image-Text Matching (ITM). Compared with existing methods that rely on the detection model to extract region-based visual features, our model uses the source image as input to conduct pre-training tasks.

\subsubsection{Masked Language Modeling}
To pre-train the model and build the mapping between language tokens and visual contents, we take the Masked Language Model (MLM) task in cross-modality domain. Specifically, we randomly mask language tokens with a probability of $0.15$, and require the model to predict the masked tokens based on other non-masked tokens and visual tokens. The learning target $\mathcal{L}_{\textrm{MLM}}$ can be formulated as
\begin{equation}
    \mathcal{L}_{\textrm{MLM}}(\theta)=-E_{(\textbf{w},I)\sim D} \log P_{\theta}(w_{m}|\textbf{w}_{\backslash m},I),
\end{equation}
where $w_m$ indicates the masked token, $\theta$ is the model parameters, and $P$ indicates the likelihood generated function.

Different from single-modality tasks with BERT where the masked tokens are only predicted from the surrounding non-masked tokens in language domain, our model can handle the cross-modality scenario where ambiguity may occur using only language modality. MLM task can encourage the model to infer the masked tokens from both language and visual tokens, which can help build the mapping between language modality and visual modality. 

\subsubsection{Image-Text Matching}
Some downstream tasks, such as image-text retrieval,  require the model to distinguish whether a sentence can well describe an image, or in other words, whether they are matched or not. To enhance the cross-modalities matching, we adopt image-text matching (ITM) task for pre-training as previous work \cite{chen2019uniter}. 
During training, we sample all image-sentence pairs provided by datasets, and consider them as positive samples. We also randomly shuffle the datasets consider the unmatched image-sentence pairs as negative samples. To prevent learning bias, we adopt the same number of positive samples and negative samples.

We apply a binary classifier on the joint embedding feature of [CLS] token to classify whether the input image and sentence are matched or not.
ITM task is driven by following loss function:
\begin{equation} \label{q_itm}
    \mathcal{L}_{\textrm{ITM}}(\theta)=-E_{(\textbf{w},I)\sim D} [y \log S_{\theta}(\textbf{w},I)+(1-y)\log(1-S_{\theta}(\textbf{w},I))],
\end{equation}
where $y\in \{0,1\}$ indicates whether the image and sentence is matched, and $S$ indicates the classification score generated function.

\subsubsection{Pixel Random Sampling}
To improve the robustness of feature learning and avoid overfitting, inspired by dropout\cite{srivastava2014dropout}, we propose to randomly sample feature pixels during pre-training. At each iteration, after extracting pixel features, we will randomly sample a part from them and feed it into Transformer. Such pixel random sampling can benefit the model training in two ways. First, it can encourage the model to learn semantic knowledge from incomplete visual input, and thus enhance the robustness. Second, it reduce the number of input elements, so that it can reduce the computation cost and accelerate the training progress. We will randomly sample a fixed number of $100$ pixels from the feature map for each input image in our experiments. Note that such pixel random sampling strategy is only applied in pre-training stage. The first reason is that random sampling in downstream tasks may lead to information missing since the fine-tuning stage only lasts for a few epochs, and another reason is that we need to make sure that inputs of downstream tasks training and testing are consistent.

\section{Experiments}

\subsection{Pre-training}

\begin{table}[t]
    \centering
    \begin{tabular}{c|c|c|c|c|c|c}
        \hline 
         \textbf{Task} & \textbf{Dataset} & \textbf{\#Imgs} & \textbf{\#Text} & \textbf{Training} & \textbf{Testing} & \textbf{Metric}  \\
         \hline \hline
         \multirow{2}{*}{Pretrain} & VG & 101K & 5.06M & train+val & - & - \\
          & COCO & 106K & 533K & train+restval & - & -\\
          \hline
         VQA & VQA2.0 & 204K & 1.1M & train+val & test-dev/test-std & VQA-score \\
         \hline
         NLVR$^2$ & NLVR$^2$ & 214K & 107K & train & dev/test & Accuracy \\
         \hline
         \multirow{2}{*}{IR \& TR} & COCO & 92K & 460K & train+restval & test & \multirow{2}{*}{Recall@1,5,10} \\
         & Flickr30K & 32K & 160K & train+restval & test & \\
         \hline
    \end{tabular}
    \caption{Statistics of different datasets, data splits and evaluation metrics used for pre-training and downstream tasks.}
    \label{tab:data_static}
\end{table}

\subsubsection{Datasets}
We pre-train our Pixel-BERT on two large-scale image-sentence datasets: MS-COCO~\cite{lin2014microsoft}, Visual Genome~\cite{krishna2017visual}. We utilize the image-level caption annotations in MS-COCO and region-level caption annotations in Visual Genome as training data for pre-training. For Visual Genome dataset, we adopt data in both \textit{train} and \textit{val} for training. For MS-COCO, we follow~\cite{karpathy2015deep} to split the whole dataset into \textit{train}, \textit{restval}, \textit{val} and \textit{test}. Since one of our downstream tasks, image-text retrieval, is conducted on MS-COCO dataset, to avoid data leak, we use \textit{train} and \textit{restval} splits for training. The statistic of training samples can be found in the first two rows of Table~\ref{tab:data_static}.

\subsubsection{Implementation Details}
During pre-training, our Pixel-BERT receives a batch of image-sentence pairs as input in each iteration. We first use the WordPiece tokenizer \cite{wu2016google} as used in BERT to split each sentence into language tokens. We use ResNet-50 as visual backbone for ablation analysis, and follow \cite{singh2018pythia,su2019vl} to adopt more powerful ResNeXt-152 to obtain better performance. We use public accessible pre-trained model on ImageNet \cite{deng2009imagenet} to initialize the parameters of the visual backbone. 
We resize the shorter edge of input images to $800$, and limit the longer edge lower than $1333$ when using ResNet-50 as visual backbone. When using ResNeXt-152, considering the GPU memory usage, we adjust the size of the shorter edge and longer edge limits to $600$ and $1000$, respectively.
As claimed in \cite{zhang2019adam}, the CNN visual backbone and Transformer may favor different kinds of the optimizer, we adopt different optimizer settings for visual backbone and Transformer. Specifically, we use SGD with learning rate $1\mathrm{e}{-2}$ and weight decay $5\mathrm{e}{-4}$ to optimize the CNN backbone, and adopt AdamW with learning rate $1\mathrm{e}{-4}$ and weight decay $1\mathrm{e}{-2}$ as Transformer optimizer. We pre-train Pixel-BERT on 64 NVIDIA Tesla V100 GPUs with the batch size 4096 samples for 40 epochs. We decay the learning rate by 10 at $25^{th}$ and $35^{th}$ epoch.

\subsection{Downstream Tasks}
We evaluate our model on several downstream vision and language tasks, including Visual Question Answering (VQA) and Natural Language for Visual Reasoning for Real (NLVR$^2)$ on VQA 2.0 \cite{goyal2017making}, NLVR$^2$ \cite{suhr2019corpus} datasets, respectively. We also conduct experiments on image-to-text and text-to-image retrieval tasks on Flickr30K \cite{xie2017aggregated} and MS-COCO \cite{lin2014microsoft} dataset. The detailed statistics including dataset splits, numbers of training/validation/testing data and evaluation metrics in all used datasets can be found in Table \ref{tab:data_static}.
We report the performance of Pixel-BERT under two different visual backbone setting in later tables, where r50 indicates ResNet-50 and x152 indicates ResNeXt-152. Since our model adopts 12-Layer Transformer as a language module, we mainly compare our experiments with other approaches under the same Transformer setting.

\begin{table}[t]
\begin{floatrow}
\capbtabbox{
\begin{tabular}{c|c|c}
    \hline 
    Model & test-dev & test-std \\
    \hline \hline
    MUTAN\cite{ben2017mutan} & 60.17 & - \\
    BUTD\cite{Anderson_2018_CVPR} & 65.32 & 65.67 \\
    \hline
    ViLBERT\cite{lu2019vilbert} & 70.55 & 70.92 \\
    VisualBERT\cite{li2019visualbert} & 70.80 & 71.00 \\
    VLBERT\cite{su2019vl} & 71.79 & 72.22 \\
    LXMERT\cite{tan2019lxmert} & 72.42 & 72.54 \\
    UNITER\cite{chen2019uniter} &  72.27 & 72.46 \\
    \hline
    Pixel-BERT (r50) & 71.35 & 71.42 \\
    Pixel-BERT (x152) & \textbf{74.45} & \textbf{74.55} \\
    \hline
\end{tabular}}{
\caption{Evaluation of Pixel-BERT with other methods on VQA.}
 \label{tab:vqa}
}
   
\capbtabbox{
 \begin{tabular}{c|c|c}
 \hline
       Model & dev & test-P\\
       \hline \hline
       Image Only\cite{suhr2019corpus} & 51.6 & 51.9 \\
       CNN+RNN\cite{suhr2019corpus} & 53.5 & 52.4 \\
       MaxEnt\cite{suhr2019corpus}  & 54.1 & 54.8 \\
       \hline
       VisualBERT\cite{li2019visualbert} & 67.4 & 67.0 \\
       LXMERT$^\dag$\cite{tan2019lxmert} & 74.9 & 74.5 \\
       UNITER$^\dag$\cite{chen2019uniter} & 75.4 & 76.0 \\
       UNITER$^\ddag$\cite{chen2019uniter} & \textbf{77.1} & \textbf{77.9} \\
       \hline
       Pixel-BERT$^\dag$ (r50) & 71.7 & 72.4 \\ 
       Pixel-BERT$^\dag$ (x152) & 76.5 & 77.2 \\
       \hline
    \end{tabular}}
    {
    \caption{Evaluation on NLVR$^2$ task.$\dag$ indicates use paired method. $\ddag$ indicates use pair-biatt method.}
    \label{tab:nlvr}
}
\end{floatrow}
\end{table}
\vspace{-6mm}

\subsubsection{Visual Question Answering}
In Visual Question and Answering (VQA) task, Pixel-BERT takes an image and a question as input and predicts an answer as output. We model it as a classification problem by learning multi-layer perception from the [CLS] token via binary cross-entropy loss. We follow the same optimizer setting as pre-training. We fine-tune the model for 18 epochs on 16 NVIDIA Tesla V100 GPUs with batch size 256. The initial learning rates are the same as pre-training, and we decay the learning rate by 10 at $12^{th}$ and $16^{th}$ epoch.

We report our experiment results on VQA task in Table~\ref{tab:vqa}. We compare our approach with recent state-of-the-art approaches. From Table~\ref{tab:vqa}, we can find that our approach with ResNet-50 as visual backbone achieves 71.35 score on test-dev split, which already outperforms ViLBERT \cite{lu2019vilbert} and VisualBERT \cite{li2019visualbert} which using more powerful visual backbone like ResNet-101 or ResNeXt-152. When equipped with ResNeXt-152 backbone, our model Pixel-BERT achieves 74.45 on test-dev split and 74.55 on test-std split, which significantly outperforms all existing works. It worth noting that this result is even higher than the performance of UNITER (Large), which using 24-Layer Transformer as a language module and get $73.40$ score on VQA \textit{test-std} split. This obvious improvement demonstrates that learning visual and language attention in image's pixel-level can benefit the visual encoder representation and enhance the visual and language embedding learning afterward.

\subsubsection{Natural Language for Visual Reasoning for Real}
Natural Language for Visual Reasoning for Real (NLVR$^2$) task requires a model to predict whether a language description is related to a given pair of images. In our model, we feed two image-language pairs into Pixel-BERT to get two embedding vectors from the [CLS] tokens, and use their concatenation to learn a classifier over ``true'' or ``false'' by cross-entropy loss. The optimizer, epoch number and learning rate settings are all the same as VQA settings explained above. And the batch size is the half of the VQA.

We evaluate NLVR$^2$ on both \textit{dev} and \textit{test-P} split. Different from pre-training tasks and other downstream tasks, NLVR$^2$ receives a pair of images at once. From the results shown in Table~\ref{tab:nlvr}, we can find that Pixel-BERT obtains 76.5 accuracy on \textit{dev} split and 77.2 accuracy on \textit{test-P} split. Our setting of composing two image-language pairs is the same as LXMERT and the ``Pair'' setting in UNITER, and from the comparison in Table~\ref{tab:nlvr} we can find that Pixel-BERT outperforms them. These results show that Pixel-BERT can be also adapted to other similar input formats.

\begin{table}[t]
    \centering
    \begin{tabular}{c|ccc|ccc}
    	\hline
         \multirow{2}{*}{Model} & \multicolumn{3}{c}{TR} & \multicolumn{3}{c}{IR} \\
         & R@1 & R@5 & R@10 & R@1 & R@5 & R@10\\
         \hline
         \hline
         VSE++\cite{faghri2017vse++} & 52.9 & 80.5 & 87.2 & 39.6 & 70.1 & 79.5 \\
         SCAN\cite{lee2018stacked} & 67.4 & 90.3 & 95.8 & 48.6 & 77.7 & 85.2 \\
         SCG\cite{shi2019knowledge} &  71.8 & 90.8 & 94.8 & 49.3 & 76.4 & 85.6 \\
         PFAN\cite{wang2019position} & 70.0 & 91.8 & 95.0 & 50.4 & 78.7 & 86.1 \\
         ViLBERT\cite{lu2019vilbert} & - & - & - & 58.2 & 84.9 & 91.5 \\
         Unicoder-VL\cite{li2019unicoder} & 86.2 & 96.3 & 99.0 & \textbf{71.5} & 90.9 & 94.9 \\
         UNITER\cite{chen2019uniter} & 84.7 & 97.1 & 99.0 & \textbf{71.5} & 91.2 & 95.2 \\
         \hline
         ours (R50) & 75.7 & 94.7 & 97.1 & 59.8 & 85.5 & 91.6  \\
         ours (X152) & \textbf{87.0} & \textbf{98.9} & \textbf{99.5} & \textbf{71.5} & \textbf{92.1} & \textbf{95.8} \\
         \hline
    \end{tabular}
    \caption{Evaluation of Pixel-BERT with other methods for image-to-text retrieval (TR) and text-to-image retrieval (IR) on Flickr30K dataset.}
    \label{tab:retrieval_flickr}
\end{table}

\begin{table}[t]
    \centering
    \begin{tabular}{c|ccc|ccc}
    	\hline
         \multirow{2}{*}{Model} &\multicolumn{3}{c}{TR} & \multicolumn{3}{c}{IR}\\
         & R@1 & R@5 & R@10 & R@1 & R@5 & R@10 \\
         \hline \hline
         \multicolumn{7}{c}{1K Test set} \\
         \hline
         VSE++\cite{faghri2017vse++} & 64.6 & 90.0 & 95.7 & 52.0 & 84.3 & 92.0\\
         SCAN\cite{lee2018stacked} & 72.7 & 94.8 & 98.4 & 58.8 & 88.4 & 94.8 \\
         SCG\cite{shi2019knowledge} & 76.6 & 96.3 & 99.2 & 61.4 & 88.9 & 95.1 \\
         PFAN\cite{wang2019position} & 76.5 & 96.3 & 99.0 & 61.6 & 89.6 & 95.2 \\
         Unicoder-VL\cite{li2019unicoder} & 84.3 & 97.3 & \textbf{99.3} & 69.7 & 93.5 & 97.2\\
         \hline
         ours (R50) & 77.8 & 95.4 & 98.2 & 64.1 & 91.0 & 96.2\\
         ours (X152) & \textbf{84.9} & \textbf{97.7} & \textbf{99.3} & \textbf{71.6} & \textbf{93.7} & \textbf{97.4}\\
         \hline \hline
         \multicolumn{7}{c}{5K Test set} \\
         \hline
         VSE++\cite{faghri2017vse++} & 41.3 & 71.1 & 81.2 & 30.3 & 59.4 & 72.4\\
         SCAN\cite{lee2018stacked} & 50.4 & 82.2 & 90.0 & 38.6 & 69.3 & 80.4 \\
         SCG\cite{shi2019knowledge} & 56.6 & 84.5 & 92.0 & 39.2 & 68.0 & 81.3 \\
         Unicoder-VL\cite{li2019unicoder} & 62.3 & 87.1 & 92.8 & 46.7 & 76.0 & 85.3\\
         UNITER\cite{chen2019uniter} & 63.3 & 87.0 & 93.1 & 48.4 & 76.7 & 85.9\\
         \hline
         Pixel-BERT (r50) & 53.4 & 80.4 & 88.5 & 41.1 & 69.7 & 80.5 \\
         Pixel-BERT (x152) & \textbf{63.6} & \textbf{87.5} & \textbf{93.6} & \textbf{50.1} & \textbf{77.6} & \textbf{86.2}\\
         \hline
    \end{tabular}
    \caption{Evaluation of Pixel-BERT with other methods for image-to-text retrieval (TR) and text-to-image retrieval (IR) on MS-COCO Dataset.}
    \label{tab:retrieval_coco}
\end{table}

\subsubsection{Image-Text Retrieval}
We consider the retrieval task as a ranking problem similar to other works \cite{chen2019uniter,li2019unicoder,su2019vl}. During training, for each image in an image-sentence pair, we use the ground-truth caption in the pair as the positive sample and randomly sample 20 unrelated captions from other pairs to make negative samples. We predict the score of whether a pair is related by a fully-connected layer on the representation of all [CLS] tokens, and apply softmax cross-entropy loss to encourage that the positive image-caption pair to obtain the highest score. We only backward the gradient on 5 negative samples with the highest loss for each image sample. Since the retrieval task is closely related to ITM task in pre-training, we only fine-tune the parameters in Transformer. We adopt AdamW with $1\mathrm{e}{-4}$ learning rate and $1\mathrm{e}{-2}$ weight decay as optimizer. We fine-tune the model on 8 NVIDIA Tesla GPUs with a batch size of 64 samples per GPU. For Flickr30K, we train 10 epochs and decay the learning rate at $6^{th}$ epoch. For MS-COCO we train 4 epochs and decay the learning rate at $2^{nd}$ epoch.

We report recall@1, 5, 10 on both text-to-image retrieval (IR) and image-to-text retrieval (TR) sub-tasks to evaluate our approach. Table~\ref{tab:retrieval_flickr} shows the 1K testing results on Flickr30K, and Tabel~\ref{tab:retrieval_coco} shows the 5-fold 1K testing results and 5K testing results on MS-COCO. We mainly compare Pixel-BERT with Unicoder-VL and UNITER which both adopt 12-layers Transformer as a language module. For image-to-text retrieval subtask, we obtain at least 0.6 performance gain on MS-COCO 1K test set, and 0.3 on MS-COCO 5K test set for recall@1. And for the text-to-image retrieval subtask, we achieve more significant results with improvements of at lease 1.9 on MS-COCO 1K testing set and 1.7 on MS-COCO 5K testing set compared with Unicoder-VL and UNITER. This is due to that the text-to-image retrieval task focuses more on the global description of an image, and our architecture can encourage the model to learn attention across language and image pixels.

\begin{table}[t]
    \centering
    \begin{tabular}{c|c|c|c|c|c|c|c}
    \hline
   \# & Visual  & Pre-traning  & Sampling  & VQA & TR& IR & NLVR$^2$ \\
    & Backbone & Tasks &Method & test-dev & val & val & dev \\
         \hline \hline
         1 & ResNet-50 & - & Random & 63.50 & 52.4 & 39.6 & 52.0\\
         2 & ResNet-50 & ITM & Random & 65.24 & 69.0 & 55.5 & 51.9 \\
         3 & ResNet-50 & MLM & Random & 71.13&67.1 & 52.7& 70.9\\
         4 & ResNet-50 & MLM+ITM & ALL &  70.84& 72.0& 57.7& 71.3\\
         5 & ResNet-50 & MLM+ITM & Random & 71.35 & 75.7& 59.8& 71.7\\
         6 & ResNext-152 & MLM+ITM & Random & 74.45 & 87.0 & 71.5 & 76.5  \\
    \hline
    \end{tabular}
    \caption{Ablation study results on VQA, Flickr30K retrieval and NLVR$^2$ downstream tasks. We evaluate the effectiveness of pre-training tasks and our proposed sampling method. The first row indicates training downstream tasks without pre-training. We report VQA score for VQA task, Recall@1 for TR and IR, and accuracy for NLVR$^2$.}
    \label{tab:ablation}
\end{table}

\subsection{Ablation Study}
We conduct ablation experiments to evaluate the effectiveness of each component of our Pixel-BERT. Since the performance of the pre-training model can not be well measured by a single metric, we use the evaluated results on downstream tasks for evaluation. The ablation study results can be found in Table~\ref{tab:ablation}. We first evaluate the effectiveness of each pre-training tasks. From the comparison of model (1) and model (2) (3), we can find that both MLM and ITM can significantly improve the performance on almost all downstream tasks. Specifically, for VQA, MLM and ITM can bring about $7.6$ and $1.6$ improvement. For the retrieval task, ITM contributes more and brings at least $13.0$ improvement on both TR and IR sub-tasks. NLVR$^2$ relies heavily on MLM task, without which the training even can not converge. The effectiveness of pre-training is consistent with the conclusions drawn in other works \cite{chen2019uniter,su2019vl}. And their combination in model (5) can further improve the performance of each task than single task. From the comparison of model (4) and model (5), we can find that our proposed randomly pixel sampling method can contribute $0.5$ score on VQA, about $2.0$  score on retrieval tasks and $0.4$ score on NLVR$^2$. This demonstrates the effectiveness of our pixel random sampling mechanism. In model (6), we follow \cite{singh2018pythia,su2019vl} to replace the visual backbone with ResNext-152, and the results show that our model with a powerful visual backbone will further accelerate the performance by a large margin.

\begin{figure}[t]
    \centering
    \includegraphics[width=\linewidth]{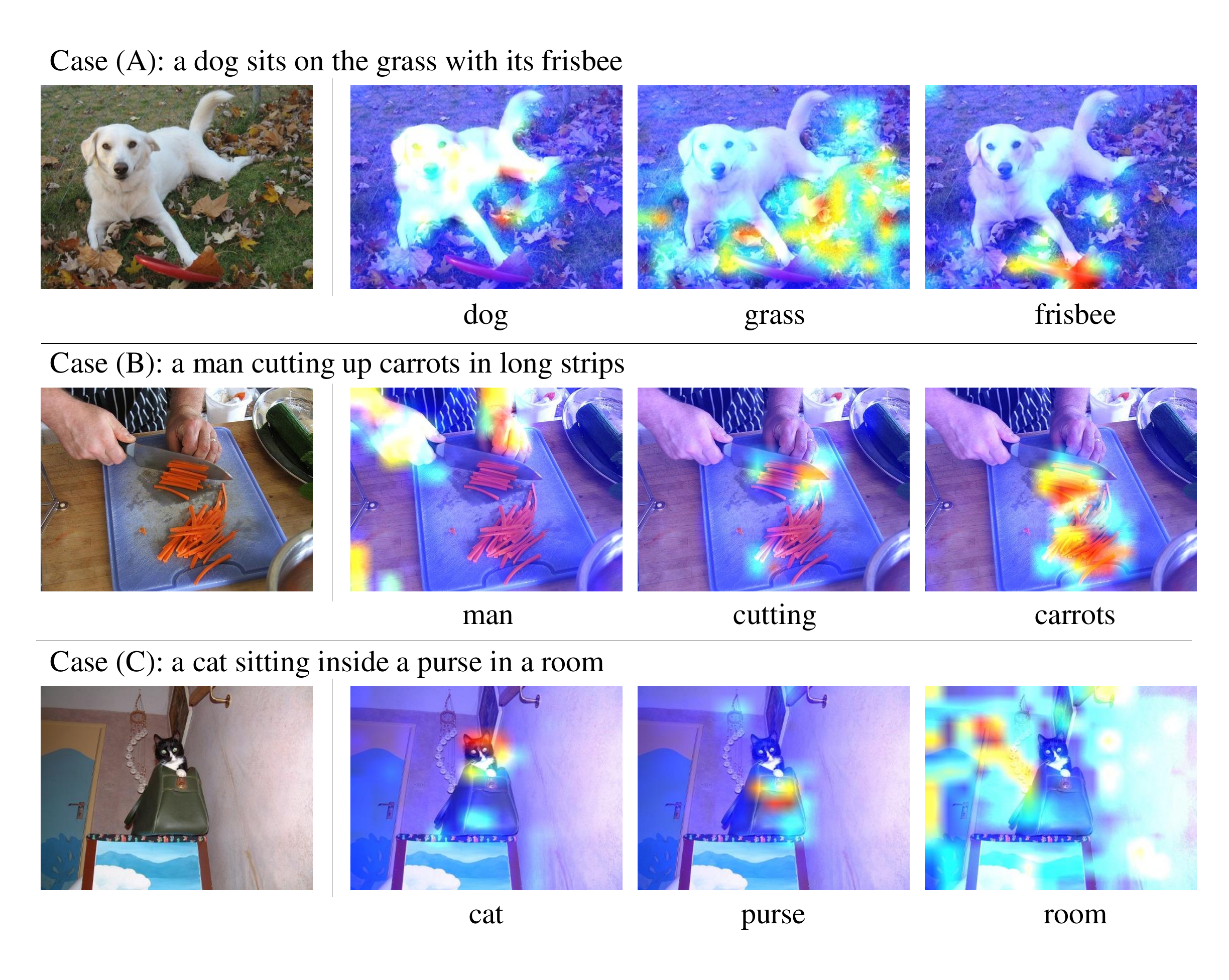}
    \caption{Visualization of attention regions extracted from the first Transformer layer of Pixel-BERT. The attention regions are extracted by using the specific token as query and pixel features as keys. Highlight areas indicate regions with high attention score.}
    \label{fig:visualization}
    \vspace{-4mm}
\end{figure}
\vspace{-5mm}

\subsection{Visualization}
To further check whether our approach Pixel-BERT can well learn the visual representation by cross-modality attention across language and pixels, we visualize some intermediate results of attention maps on examples from MS-COCO \textit{val} set.
The visualization results can be found in Fig.~\ref{fig:visualization}. From the result of Case (A), we can see that the response areas of token ``dog'', ``grass'' and ``frisbee'' are actually distributed on the correct region. For Case (B), we can find that although ``cutting'' is a verb, it can attend to the most related region in which the action of ``cutting'' is performed with a knife. From Case (C), we find that the token ``room'' can attend to the correct region in the image, which is difficult to be represented by a bounding box. Although we did not apply any spatial supervision (e.g., bounding box annotations) to guide the attention learning, the results from Fig.~\ref{fig:visualization} show that with well-defined tasks, our Pixel-BERT can well learn the visual representation in region level with cross-modality learning. This result also brings a lot of possibilities for further research to study whether cross-modality learning can help the semantic understanding of visual information conversely.




\section{Conclusion and Discussion}
Pre-training mechanism has shown its effectiveness in vision and language domain.
In this paper, we discuss the visual embedding method that is commonly used in existing works and aim to solve the limitation of region-based visual representation. 
We propose CNN-based Visual Encoder and combine it with multi-modal Transformers to construct Pixel-BERT in an end-to-end manner and build more accurate and more thorough embedding between visual and linguistic contents in pixel and text level.
We use pixels of an image as input and apply a random pixel sampling mechanism for the robustness of visual embedding learning.
We build a pre-training model based on Pixel-BERT to learn a universal visual and language embedding on Visual Genome dataset and MS-COCO dataset. Masked language model and image-text matching are two tasks designed for pre-training. 
We conduct downstream vision and language tasks with our pre-trained model and achieve the best performances in most tasks, including VQA, NLVR$^2$, image-to-text retrieval and text-to-image retrieval.

Without the restriction of annotated bounding boxes, our pre-trained model with Pixel-BERT can provide much stronger representation for both images and sentences with a larger image-sentence pair dataset. We will consider to pre-train our model on Conceptual Caption Dataset \cite{sharma2018conceptual} to further optimize the visual and language embedding.
Masked visual prediction is proposed in some works. For example, \cite{chen2019uniter} proposed three types of masked region modeling for the prediction of the masked region. In this paper, we use a random pixel sampling mechanism to replace this part due to the difficulty of pixel reconstruction compared with regions. In the future, we will research on how to design and combine self-supervised tasks for visual contents in our current approach.

%
%
\bibliographystyle{splncs04}
\bibliography{egbib}
\end{document}